\documentclass{article}
\usepackage{arxiv}
\usepackage[utf8]{inputenc}
\usepackage[T1]{fontenc}
\usepackage{hyperref}
\usepackage{url}
\usepackage{booktabs}
\usepackage{amsmath,amssymb,amsfonts}
\usepackage{nicefrac}
\usepackage{microtype}
\usepackage{graphicx}
\usepackage[numbers,sort&compress]{natbib}
\usepackage{doi}
\usepackage{cleveref}
\usepackage{multirow}
\usepackage{float}
\usepackage{textcomp}
\usepackage{xcolor}
\usepackage{algorithm}
\usepackage{algorithmic}
\usepackage{bm}
\usepackage{authblk}

\title{Whispering Wisely: Smart Subset Selection for ASR Fine-Tuning}
\date{}

\author[1,2]{%
  \href{https://orcid.org/0000-0001-5180-4541}{Roi Yozevitch}%
  \thanks{Corresponding author: \texttt{Yozevitch@gmail.com}.}%
}
\author[1]{%
  \href{https://orcid.org/0000-0002-7673-6511}{Or Haim Anidjar}%
}

\affil[1]{Department of Computer and Software Engineering,
Ariel University, Israel}
\affil[2]{Department of Electrical and Electronic Engineering,
Ariel University, Israel}

\begin{document}
\maketitle

\begin{abstract}
Automatic Speech Recognition (ASR) models such as OpenAI’s Whisper have substantially improved speech recognition quality but remain expensive to fine-tune, especially when large amounts of transcribed audio are available. This makes it challenging for many practitioners to adapt these models to new languages or domains. In this work, we introduce an embedding-based subset selection approach for selecting smaller training sets for ASR fine-tuning. Using Whisper’s pretrained encoder, we map each audio sample into a fixed-dimensional embedding space and apply a greedy k-center algorithm to select diverse and representative subsets of the training data. We evaluate this approach on both Hebrew and Danish, showing that k-center selection often improves upon random sampling, particularly at smaller subset sizes. An analysis of lexical diversity reveals that k-center subsets capture a broader range of linguistic content than random subsets of the same size, which may help explain their observed effectiveness. Overall, our results demonstrate that embedding-based subset selection provides a practical strategy for ASR subset selection in the two evaluated languages, with the clearest recognition benefits observed at smaller sample budgets.
\end{abstract}

\keywords{Automatic Speech Recognition \and Whisper \and Embedding-Based Subset Selection \and K-Center Algorithm}

\section{Introduction}\label{sec:intro}
Automatic Speech Recognition (ASR) is the technology that converts spoken language into text. Over decades of development, ASR systems for major languages like English have reached near-human level accuracy \cite{liu2024exploration}, powering voice assistants, transcription services, and other everyday applications \cite{loubser2024end, khan2024mser}. But among the world's 7,000+ languages, most still lack the large training datasets and resources required to build high-quality ASR systems. Low-resource languages (those with limited audio-text data available \cite{anidjar2025transformer, anidjar2023crossing, anidjar2024whisper}) often have much higher error rates and unreliable performance \cite{adaptive_speech, zhao2022improving}. For example, Hebrew is an under-resourced language with 9 million speakers. In the past, it lacked many ASR resources, which caused older models to have high error rates \cite{silbererror} and made it impractical for daily use. This gap has led to a growing focus on developing ASR solutions for low-resource languages \cite{SMILE}, aiming to deliver quality speech recognition to all communities.  

One promising development is the rise of large-scale pre-trained speech models. OpenAI's Whisper model \cite{whisper} is a prime example: It was trained on 680,000 hours of multilingual speech. Unlike traditional ASR models trained on a single language, Whisper's multilingual training (reportedly covering around 100 languages) enables it to generalize across different languages and achieve state-of-the-art performance out of the box \cite{whisper}. Such a model offers a huge leap forward for low-resource languages -- in principle, one can take the pre-trained Whisper and fine-tune it on a small dataset of a target low-resource language to get a high-accuracy recognizer \cite{ft_whisper_low_resource_aaa}.

Whisper's different versions range from 39 million to 1.5 billion parameters, and updating all weights for a new language or domain requires substantial GPU time and energy. Compared to training a smaller model from scratch, fine-tuning a billion-parameter model can take a lot more time and memory -- often running for days even on high-end hardware \cite{energy}. The cost is not just time: it also translates to financial cost (for cloud compute or electricity) and environmental impact \cite{carbon_footprint}.

Recent studies have quantified the energy used in training modern deep learning models to be huge -- for example, training a single large neural NLP model with an extensive hyperparameter search was estimated to emit over 626,000 pounds of CO$_2$ \cite{energy}, which is about five times the lifetime emissions of an average car. Similarly, \cite{bloom} estimated that training the 176B-parameter BLOOM model resulted in emissions on the order of tens of tonnes of CO$_2$. While fine-tuning on a smaller dataset would not reach that extreme,  the point remains that large model training is resource intensive and not accessible to everyone. In the context of ASR, researchers working on less common languages often have a limited computing infrastructure \cite{asrafrican}. Even well-resourced organizations have incentives to reduce training time and energy usage, since faster training means lower cost. Thus, the question arises: How can one adapt massive ASR models like Whisper to new languages or tasks efficiently, without weeks of GPU time?

Hebrew ASR \cite{varod2021cross} is a concrete example of this challenge. Until recently, there was no large-scale Hebrew speech data to train advanced models. This changed with the introduction of the Ivrit.ai \cite{marmor2023ivritai}, which offers over 3,300 hours of transcribed Hebrew speech from diverse speakers. However, using all 3,300 hours in practice is easier said than done -- fine-tuning on thousands of hours of audio is computationally expensive for most research labs or companies. Training on the full dataset would take several days on a multi-GPU server, which is not practical outside of large industry labs. In many cases, the available time and computing budget simply do not permit using the entire dataset. Thus, practitioners might have to subsample the data or use a smaller model, trading away some accuracy because they cannot afford the full training process. Even if one has the resources, training on the whole dataset every time is inefficient, especially if a lighter approach could achieve almost the same result. The efficiency of the fine-tuning process thus becomes crucial in making high-quality ASR accessible in smaller-scale settings. 

An important observation is that not all training examples are equally valuable to the model \cite{mohiuddin2022data_aaa, grad-match_aaa}. Large speech datasets often contain redundant information, many recordings cover the same vocabulary or acoustic patterns, so training on all of them gives diminishing returns \cite{submodular}. It is well-documented that increasing the size of training sets yields smaller and smaller improvements over time \cite{scale_laws,dl_scale}, illustrating the classic diminishing returns seen in machine learning. Additional data might be noisy, or simply repeat what is already covered, and thus contribute little new meaning. This phenomenon is partly explained by the low entropy and redundancy inherent in natural languages: languages are predictable due to repeated words, common phrases, and limited structural variation. As a result, larger speech datasets inevitably contain plenty of repetitive information. Crucially, this means that a significant portion of a huge dataset can be removed with minimal impact on final accuracy, if one manages to keep the "right" data. Previous works in ASR show this: having too much redundant data wastes computation, so the goal should be to select a small informative and representative subset that preserves the performance of using the full data \cite{rangappa2025efficient, kothawade2021ditto}.

This idea of smart subset selection has been explored in other domains with promising results. In computer vision, for example, core-set selection and active learning methods try to choose a subset of images such that training a model on that subset gives almost the same accuracy as training on the full dataset. \cite{alcnn} showed that by selecting the most representative samples, the performance of the model trained on the selected data remains competitive. In deep learning more generally, data selection techniques have shown that large parts of training data can be pruned away without harming accuracy. \cite{proxy} present a selection via proxy approach where a small proxy model quickly estimates which data points are more useful. Their method was able to remove 50\% of CIFAR-10 training images with no loss in final accuracy of the full model, achieving a $1.6\times$ faster training time. In the field of machine translation, \cite{mohiuddin2022data_aaa} reported that by fine-tuning on carefully chosen subsets of data, the model only converged $2\times$ faster (requiring about half of the updates) and even improved translation quality compared to using all data. These examples prove a powerful point: more data is not always better. A smaller, wisely selected dataset can often give you the same (or better) results at lower cost.

In this work, we tackle the problem of efficient ASR fine-tuning through subset selection, using Hebrew and Danish as case studies. We assume a scenario where a large pre-trained model (like Whisper) is available, and we have a large pool of transcribed audio (such as the Ivrit.ai dataset and comparable Danish resources) but limited resources \cite{anidjar2024extending} to train on it fully. We propose a practical embedding-based subset selection method. In short, we compute vector embeddings from our samples and use a clustering/diversity criterion to select a representative subset. This has a one-time cost, but after that, training proceeds on the subset with no additional overhead. Because the method yields similar improvements for both Hebrew and Danish (which are languages with very different phonological and data characteristics), the experiments provide observations in two language settings, aligning with the multilingual nature of Whisper’s encoder.

\subsection{Contributions}
Our contributions are summarized as follows:

\begin{enumerate}
    \item \textbf{Efficient Subset Selection Method} We introduce an embedding-based subset selection for ASR fine-tuning. We represent each audio sample as an embedding in a feature space and then select a diverse, representative subset. Our method requires only a one-time embedding computation cost, reduces the amount of training data needed, and adds no overhead during training, making it highly practical for real-world use. Embedding extraction and subset selection are performed before fine-tuning and do not require selection updates during training.
    
    \item \textbf{Linguistic Diversity Preservation} Through quantitative analysis, we show that embedding-based k-center selection yields higher unique-word counts than random sampling in the evaluated subsets, particularly for Hebrew. These counts describe vocabulary coverage; they do not directly measure phonetic variation.
    
    \item \textbf{Evaluation in Two Languages} The selection procedure requires no language-specific selection rules and was evaluated on Hebrew and Danish datasets. These experiments provide evidence in two language settings rather than establishing generalization to arbitrary languages or datasets.
 
    \item \textbf{Potential Annotation Applications} Since selection uses audio embeddings, the procedure could potentially guide which recordings to transcribe. Annotation savings and recognition performance under such a labeling protocol were not evaluated in this study.
\end{enumerate}

\subsection{Paper Structure}
The rest of the paper is structured as follows: 

Section~\ref{sec:related_work} reviews prior research on computationally efficient training for large speech models, with emphasis on data pruning and diversity-driven subset selection relevant to ASR.
Section~\ref{sec:dataset} presents the Hebrew and Danish corpora used in our experiments, describing their linguistic and acoustic characteristics.
Section~\ref{sec:approach} presents the embedding-based k-center subset selection pipeline built on Whisper encoder representations and explains how we fine-tune models on reduced training sets.
Section~\ref{sec:results} reports ASR performance across subset sizes, distance metrics, and baselines, and includes an analysis of the lexical diversity captured by the selected subsets.
Section~\ref{sec:conclusion} summarizes the main findings and discusses the practical implications of embedding-based data selection for efficient, resource-constrained ASR fine-tuning.

The abbreviations used throughout the paper are listed in Table~\ref{tab:abbr}.

\begin{table}[H]
\centering
\caption{List of abbreviations}
\label{tab:abbr}
\begin{tabular}{ll}
\hline
\textbf{Abbreviation} & \textbf{Meaning} \\
\hline
ASR & Automatic Speech Recognition \\
CER & Character Error Rate \\
WER & Word Error Rate \\
NLP & Natural Language Processing \\
DDP & Dynamic Data Pruning \\
LM & Language Model \\
PCA & Principal Component Analysis \\
\hline
\end{tabular}
\end{table}

\section{Related Work}
\label{sec:related_work}
The past few years have seen ASR models grow in size and complexity.
Large-scale pre-trained models like Whisper \cite{whisper},
wav2vec 2.0 \cite{wav2vec}, and HuBERT \cite{hubert} achieve impressive
accuracy, but training or fine-tuning these models for specific languages
or domains is computationally expensive. Beyond ASR, neural speech
representations have also been studied for multilingual spoken language
recognition using TDNN-based architectures and temporal pooling
\cite{anidjar2025enhancing}, although that work addresses language
identification rather than training-data selection.

Low-resource organizations often cannot afford to train such massive
models from scratch on domain-specific data. As a result, researchers
are exploring techniques to improve training efficiency, either by
reducing the amount of data required or by lowering computing demands.

We divide related work into two main categories: utility-based methods
(Section~\ref{sec:sec21}), which rely on training feedback (e.g. loss,
gradients, uncertainty scores) to identify the most valuable samples;
and coverage-based methods (Section~\ref{sec:sec22}), which select a
representative, diverse subset based on data distribution or embeddings,
independent of model training signals.

\subsection{Data Pruning and Sample Selection Strategies in ASR}\label{sec:sec21}

\begin{figure*}[htbp]
    \centering
    \includegraphics[width=\linewidth]{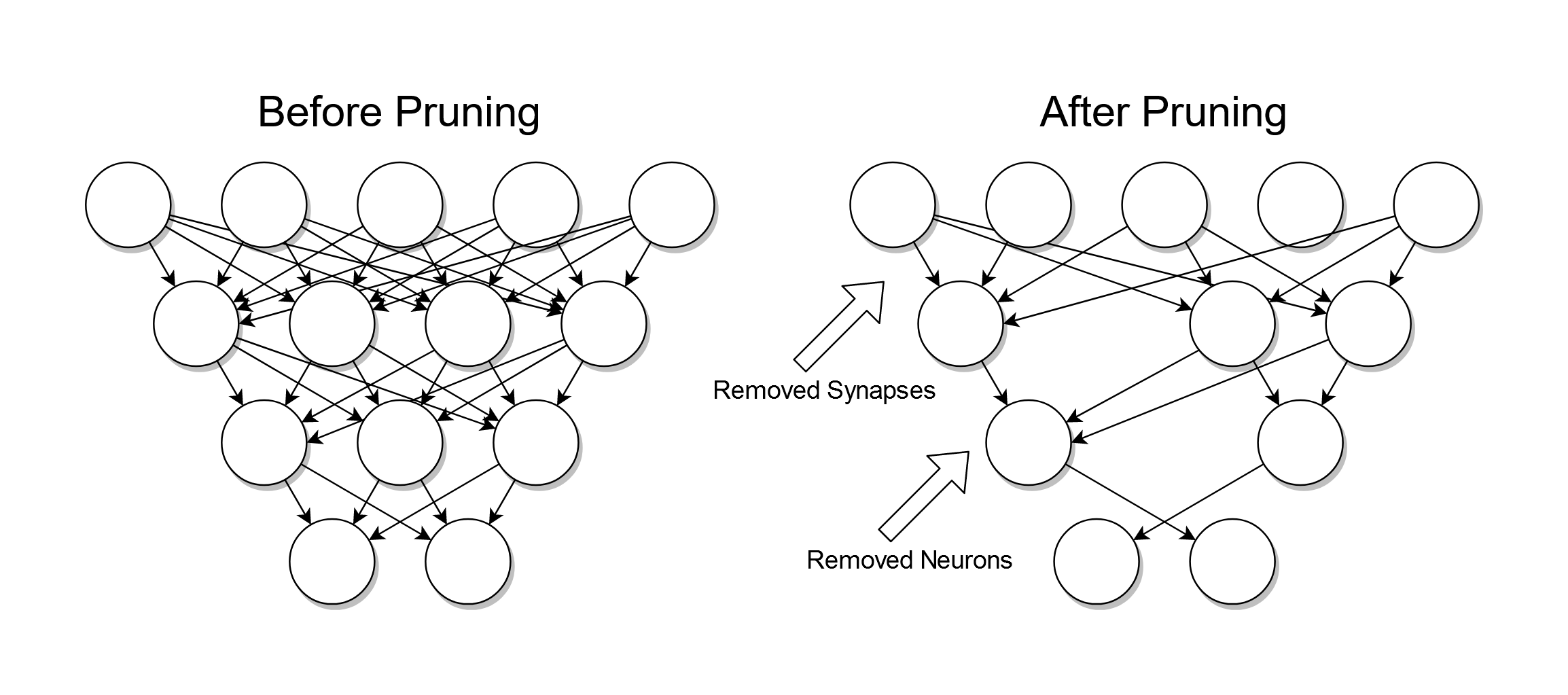}
    \caption{An illustration of the data pruning process, employed in this paper (taken from \href{https://commons.wikimedia.org/wiki/File:Before_after_pruning.png}{Wikipedia Commons}).}
    \label{fig:data_pruninng}
\end{figure*}

One main line of work in this study is data pruning (Figure~\ref{fig:data_pruninng}), that is, selecting a smaller training subset that still maintains high performance. The goal is to remove redundant or less informative examples and save training time. Several methods have been proposed in this area, each with different ways to identify which training samples are most useful:

\cite{cowerage} introduced an algorithm called COWERAGE that selects a subset of training examples based on their early-stage WER during fine-tuning. The idea is to identify "hard" samples (those with high WER in the first few epochs) and make sure the subset covers those challenging segments. By prioritizing these harder, more informative samples (which also tend to be more phonetically diverse), the model can generalize better. For example, COWERAGE achieved up to a 17\% relative WER improvement over random data sampling in their experiments. The main drawback is the extra overhead: COWERAGE requires running a partial training run (or a few epochs of fine-tuning) on the full data to estimate per-sample WER, effectively adding a second phase to the training pipeline.

\cite{ddpasr} took a dynamic approach by continuously updating the training set as learning progresses. Instead of choosing a fixed subset beforehand, their Dynamic Data Pruning (DDP) method prunes less useful data on the fly. At regular intervals (e.g., each epoch or batch), the algorithm computes a pruning score for each sample (they used the instantaneous loss of each training example as the score), and drops the batches or instances with the lowest utility. This is a dynamic selection: an example not used in one epoch could be reintroduced later if it becomes informative under the evolving model. This adaptive pruning strategy was shown to reduce training requirements without hurting accuracy. On the Librispeech \cite{librispeech} ASR dataset, training on only about 70\% of the data (the hardest samples by loss) gave the same WER as training on the full 100\% of the data. In other words, the least informative 30\% of examples could be skipped with no performance drop, which means a lot of redundancy in the dataset. This translates to about 1.6$\times$ faster training time in their setup. The benefit of DDP-ASR is that it never discards any data; the pruning decision is made based on the model’s current learning status. However, there is an added complexity in the training. 

Another approach is to analyze training dynamics to identify valuable vs. redundant examples. \cite{dynuncertain} presented a large-scale dataset pruning method using dynamic uncertainty measures, inspired by the concept of data cartography. By tracking each training example’s prediction confidence and variability across epochs, one can find which examples are consistently easy (high confidence, low loss) and which are ambiguous or hard (low confidence or fluctuating predictions). Their results showed that filtering out the most redundant or consistently easy samples can reduce the dataset without hurting accuracy. In fact, they were able to retain only about one-third of the data and still match the full-dataset performance in some cases. This means many training examples contribute little once the model has seen enough similar instances. Such methods, however, require monitoring the model’s behavior over multiple epochs or training runs. Implementing them can be tricky, since you need to log per-sample metrics and possibly restart training with the pruned set once identified.

\cite{rangappa2025efficient} proposed a multi-stage data selection pipeline to adapt ASR models to new domains using unlabeled data. They generated pseudo-transcripts for a large in-domain audio pool using Whisper \cite{whisper} and a Zipformer \cite{zipformer}, then applied iterative filtering based on predicted word error rate (WER), cross-model agreement (character error rate across multiple ASR outputs), and the presence of domain-specific named entities. The result is a high-quality subset of training segments: for example, fine-tuning on only 100 hours (about 1.4\% of their original dataset) of pseudo-labeled call-center speech data  achieved a similar WER compared to fine-tuning on the entire dataset. A key strength of this approach is that it combines multiple selection criteria; it eliminates segments that are likely to be transcribed incorrectly (high predicted WER) or where different ASR models disagree (high CER), while ensuring important keywords/entities are covered. On the downside, the pipeline’s complexity is a consideration. It relies on training auxiliary models and running multiple ASR systems to evaluate each candidate segment.

Unlike the above methods, our approach is an offline preprocessing step that does not add extra complexity or delay during model training. It has a one-time cost to compute embeddings and run a subset selection algorithm. The trade-off is that an offline method lacks the real-time \cite{kfir2025real} feedback of dynamic approaches. But, as we show, a well-chosen static subset can get competitive WER, making it an alternative when integrating complex pruning algorithms into the training pipeline is not desirable. 

\subsection{Coverage and Diversity-Based Subset Selection}\label{sec:sec22}

Another line of work treats the subset selection problem as one of maximizing coverage or diversity of the full dataset. The idea is that a smaller subset can still train a good model if it represents the diversity of the whole data distribution in the full set. Several methods were proposed based on this idea:

\cite{alcnn} took an approach of choosing a subset such that every data point in the full set is near some selected point in feature space. They used a greedy farthest-point algorithm (iteratively picking the data point that maximizes the minimum distance to the already selected set) to construct a core-set. In image classification experiments, this method allowed training on significantly fewer labeled images while maintaining almost the same accuracy as using all data. The strength of this farthest-first selection is that it explicitly seeks out outliers and under-represented regions in the feature space, ensuring good coverage of all clusters of examples. The downside is computational cost: naive implementation requires
computing distances between all points, which can be expensive for very large datasets.

Another direction involves submodular selection: Submodular set functions provide a way to balance diversity and representativeness. A set function is submodular if it exhibits diminishing returns (adding an item to a smaller set yields a larger gain than adding it to a bigger set). Researchers have designed submodular utility functions for subset selection in machine learning tasks. For example, a facility-location function can be used where each data point has a "facility" cost and selecting a point gives a benefit to its nearby points; this encourages choosing points that cover many others \cite{submodlib}. In the context of speech data, \cite{submodular} applied submodular selection to large-scale ASR training data and showed that by defining an appropriate submodular utility function (e.g., based on likelihood or clustering criteria), one can remove a substantial portion of training examples with minimal performance loss. 

\cite{kothawade2021ditto} tackled the problem of ASR performance disparities across different speaker accents by selecting the data that best improves a target accent’s accuracy. Their system, DITTO, uses submodular mutual information selection strategies to find a subset of examples that are most informative for a given accent within a fixed labeling budget. In practice, you provide a small set of example samples in the target accent (as a "query" set), and DITTO ranks a large pool of candidate audio by how well they cover the accent. This targeted selection was very data-efficient: on benchmark accent corpora, DITTO achieved the same WER reduction with only 20–33\% of the data that other methods required, i.e. it was 3–5× more label-efficient than random or acoustically-diverse selection baseline. By focusing on accent-specific informative samples, they significantly reduced the annotation needed while mitigating ASR bias against those accents.  

\cite{discrete_represent} presented a simple unsupervised data selection method to improve self-supervised ASR pre-training on limited or domain-specific data. They proposed to select acoustically relevant parts of a big corpus by using discrete speech units from a pre-trained model. In their approach, they transform speech into a text-like representation by passing the raw audio through a pre-trained encoder. Then, following a contrastive data selection strategy, they train language models (LMs) on the discrete token sequences. By comparing these LMs, they assign each candidate sample a domain-relevance score. Samples most similar to the target domain’s token distribution are kept, while others are filtered out. This requires no transcripts or human labels – it’s entirely unsupervised – and works well. They show that pre-training a model on a 6\% subset of a massive audio pool can outperform pre-training on the full set of data. They showed that using only 6\% of the data resulted in an 11.8\% relative WER improvement. 

In our diversity-based approach, we use Whisper's pretrained embedding space to define the geometry. Prior diversity selection methods operated on raw features or learned representations from scratch; we instead use Whisper’s rich encoder representations as a high-level feature space that already captures acoustic and linguistic patterns relevant to ASR. The idea is that if two audio clips have very similar Whisper encoder embeddings \cite{anidjar2021hybrid, anidjar2023speech, anidjar2024harnessing, anidjar2020thousand, anidjar2021thousand}, they likely contain redundant information for the ASR task (perhaps they have the same phonetic content). In contrast, if the embedding of a sample is far from others, it might represent a unique speaking style or a rarer linguistic content. By computing the embeddings of all training samples and then selecting a subset that maximizes coverage in this space, we inherently prioritize diversity in terms of the model's learned features. This can be seen as a simplified form of facility-location: Each selected sample "covers" its neighboring samples in the embedding space up to some radius. We do not manually craft a complex objective function; instead, we rely on distance metrics to guide a greedy selection of diverse points.

\section{Dataset}\label{sec:dataset}

We run all experiments on the Ivrit.ai Hebrew speech corpus \cite{marmor2023ivritai}, the largest public resource for Hebrew ASR. Ivrit.ai contains over 3,300 hours of Hebrew speech together with their transcription. The dataset includes recordings from various real-world sources such as podcasts and online lectures. These recordings reflect different speaking styles, accents and acoustic environments, making it a comprehensive benchmark for Hebrew ASR.

However, working with a Hebrew ASR corpus has its own challenges. First, Hebrew is a morphologically rich and Semitic language with complex grammar and many inflectional paradigms. Unlike English, Hebrew does not have vowels in the orthography, which creates ambiguity that makes speech recognition significantly more difficult \cite{hebDB}. Second, the Ivrit.ai dataset consists of data from varied sources, so it includes a wide range of audio quality and acoustic conditions. The variability in recording conditions, such as background noise, microphone differences, informal speaking styles, makes it hard for ASR models to generalize across these conditions.

From this corpus, we randomly sampled 20,000 audio clips (approximately 28 hours of audio) together with their transcriptions to serve as our initial training dataset. We chose this number for the following reason: A significantly smaller training dataset (one with a few thousand samples) would be computationally manageable without pruning, so there would be no real need for subset selection. Conversely, using a very large dataset (e.g., 50,000 or more samples) would mean that relatively large subsets (e.g., 70\% of the data) would work almost as well as the whole dataset, making it hard to see the effects of subset selection. Therefore, 20,000 samples hit the sweet spot; it is big enough to show clear trade-offs when we pick subsets, but still small enough that we can fine-tune Whisper on all of it to get a solid baseline to compare with.

In addition to the training data, we put aside a separate evaluation dataset of 4,000 samples. We evenly split these samples into two subsets of 2,000 samples each: a validation set used to track the training process and prevent overfitting; and a test set for the final evaluation. We made sure that these evaluation sets did not overlap with or leak into the main training set. 
To align with Whisper's model requirements, we first converted all audio samples to 16kHz single-channel format. Next, we cleaned up the transcripts by removing punctuation and special characters.

In addition to Hebrew, we evaluate our subset selection method on Danish using a 20,000-sample subset of the CoRal-v2 corpus \cite{coral2024}. CoRal-v2 is a large publicly available Danish speech resource that provides several hundred hours of transcribed read-aloud speech from speakers across Denmark. From this corpus, we select 20,000 samples (approximately 34 hours of audio) and, as with Hebrew, we applied the same preprocessing and put aside validation and test sets following the same ratios. Including Danish complements our Hebrew experiments by testing the method on a typologically different language. This allows us to assess whether the subset selection approach generalizes across languages with distinct phonological structures.

\begin{figure}[htbp]
    \centering
    \includegraphics[width=\columnwidth]{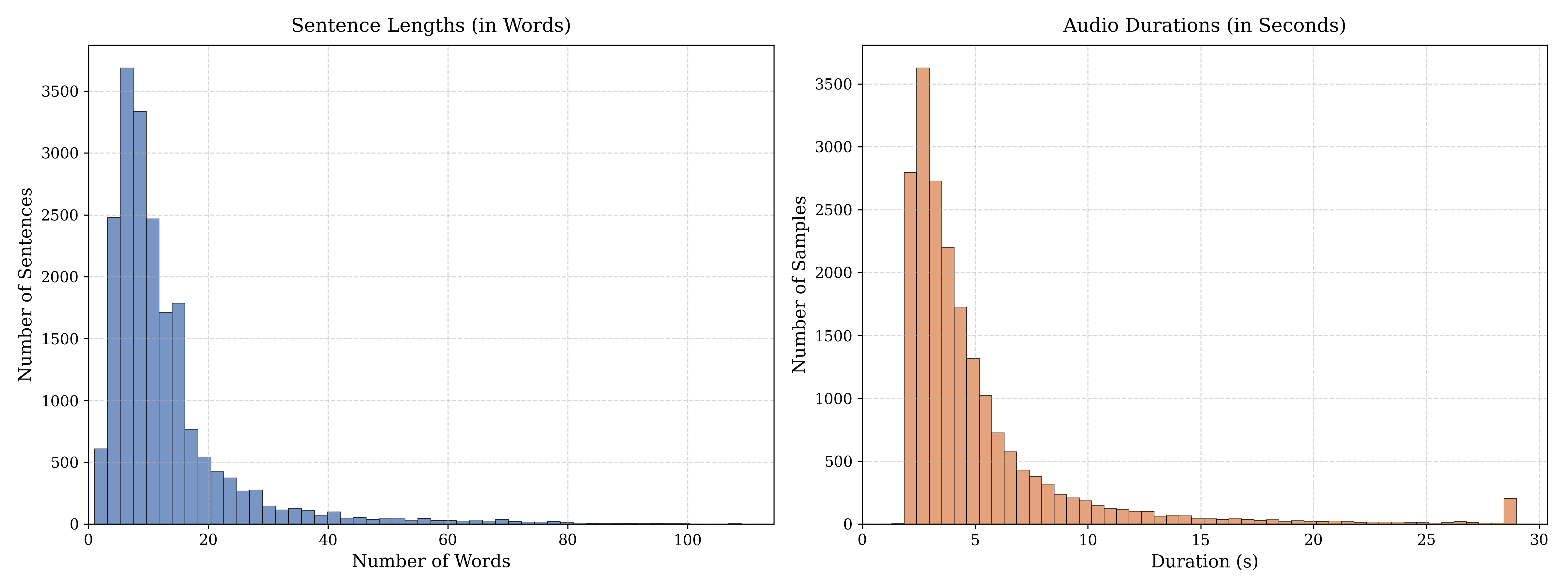}
    \vspace{0.3em} 
    
    \includegraphics[width=\columnwidth]{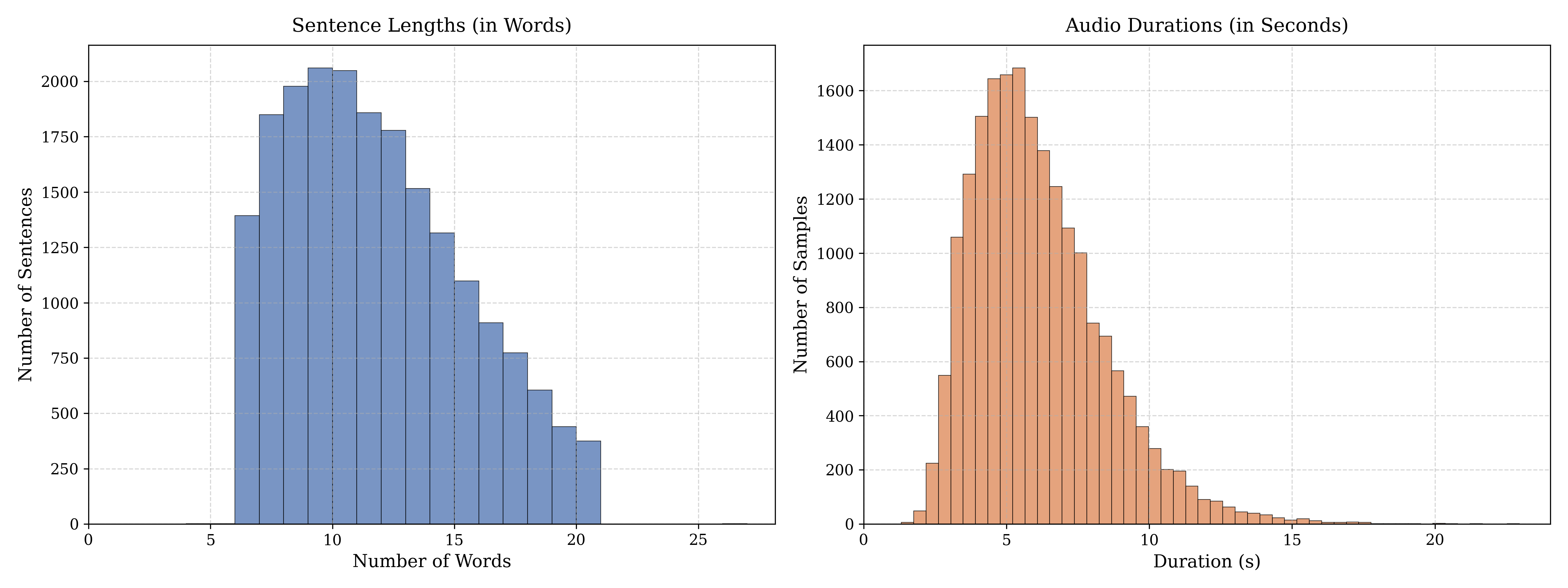}

    \caption{Distributions of sentence lengths and audio durations for the Hebrew dataset (top) and the Danish dataset (bottom), each based on 20,000 samples.}
    \label{fig:data_distributions}
\end{figure}

Figure \ref{fig:data_distributions} shows histograms that illustrate statistical properties of our training data. For Hebrew, the distribution of sentence lengths shows that most sentences contain between 5 and 20 words, while audio durations peak around 3–4 seconds and gradually decline towards the 15 seconds range. In contrast, the Danish subset has a much narrower span: sentence lengths are concentrated between roughly 8 and 12 words, and audio durations are more tightly grouped around 4–8 seconds. Taken together, the two datasets provide complementary linguistic and acoustic characteristics, giving us a broad and representative setting for evaluating our subset selection method. 

\section{Our Approach}\label{sec:approach}

\begin{figure}[htbp]
    \centering
    \includegraphics[width=\columnwidth]{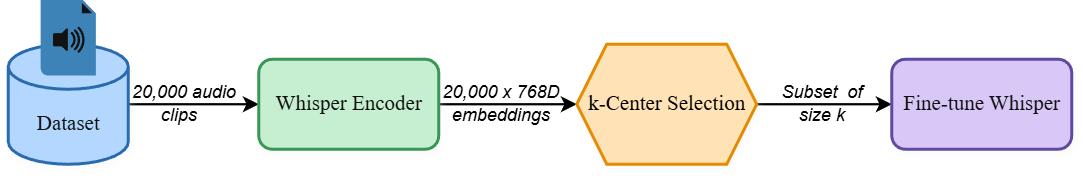} 
    \caption{An overview of our embedding-based subset selection pipeline. Starting with 20,000 clips, we extract 768-dimensional embeddings via Whisper's encoder, apply a greedy $k$-center selection for a subset of size $K$, and use these to fine-tune the model.}
    \label{fig:diagram}
\end{figure}

Our goal is to show that high-quality ASR models can be fine-tuned on much smaller data sets and still achieve competitive accuracy. Training large-scale ASR models such as Whisper on thousands of hours of speech data requires substantial computational resources and training time, resources that are often unavailable to smaller research groups, start-ups, or organizations working with under-resourced languages. To bridge this gap, we propose a practical way to select smaller yet highly representative subsets from large speech corpora, so that models such as Whisper can be fine-tuned with reduced computational overhead and minimal accuracy loss. We follow the four-step pipeline depicted in Figure~\ref{fig:diagram}. 

Subset selection is performed as an offline preprocessing step. It reduces the number of utterances supplied to fine-tuning, but smaller sample counts do not by themselves establish lower total runtime or peak memory requirements.

\subsection{Data Pool Creation and Preprocessing}
For both Hebrew and Danish, we created an initial data pool by randomly sampling 20,000 audio–text pairs from each corpus. We applied the same preprocessing steps described in Section ~\ref{sec:dataset} to both languages, including transcript normalization and resampling all audio to a 16 kHz sampling rate. This ensured consistent inputs in the two datasets and provided a uniform starting point for the embedding and subset selection steps that follow.

\subsection{Embedding Representation via Whisper's Encoder}
We base our subset selection strategy on Whisper encoder embeddings, which provide high-level speech representations. Whisper’s encoder, consisting of multiple convolutional and transformer layers, transforms raw audio waveforms into a high-dimensional space that captures acoustic, phonetic and linguistic information. Since the encoder is trained on hundreds of thousands of hours of multilingual speech, these embeddings inherently reflect meaningful properties relevant for speech recognition tasks. To obtain a single embedding vector per sample, we pass each preprocessed audio sample through Whisper’s encoder (frozen, i.e., not updating its parameters). The encoder produces a sequence of hidden states for each segment of the input audio. We applied mean pooling across the time dimension to convert these variable-length sequences into fixed-size, 768-dimensional embedding vectors, which represent the acoustic and linguistic content of the audio clip. These embeddings are the basis for our subset selection step, allowing us to use Whisper’s pre-trained knowledge.

\subsection{Representative Subset Selection Using Greedy k-Center Algorithm}

Once the embeddings are computed, the next step is to select a smaller subset of audio samples that captures the original data distribution as much as possible. Our core idea is that the subsets chosen based on their representativeness and diversity in the embedding space can match the performance of training on much larger, randomly sampled datasets.

\begin{figure}[htbp]
    \centering
    \includegraphics[width=1.0\columnwidth]{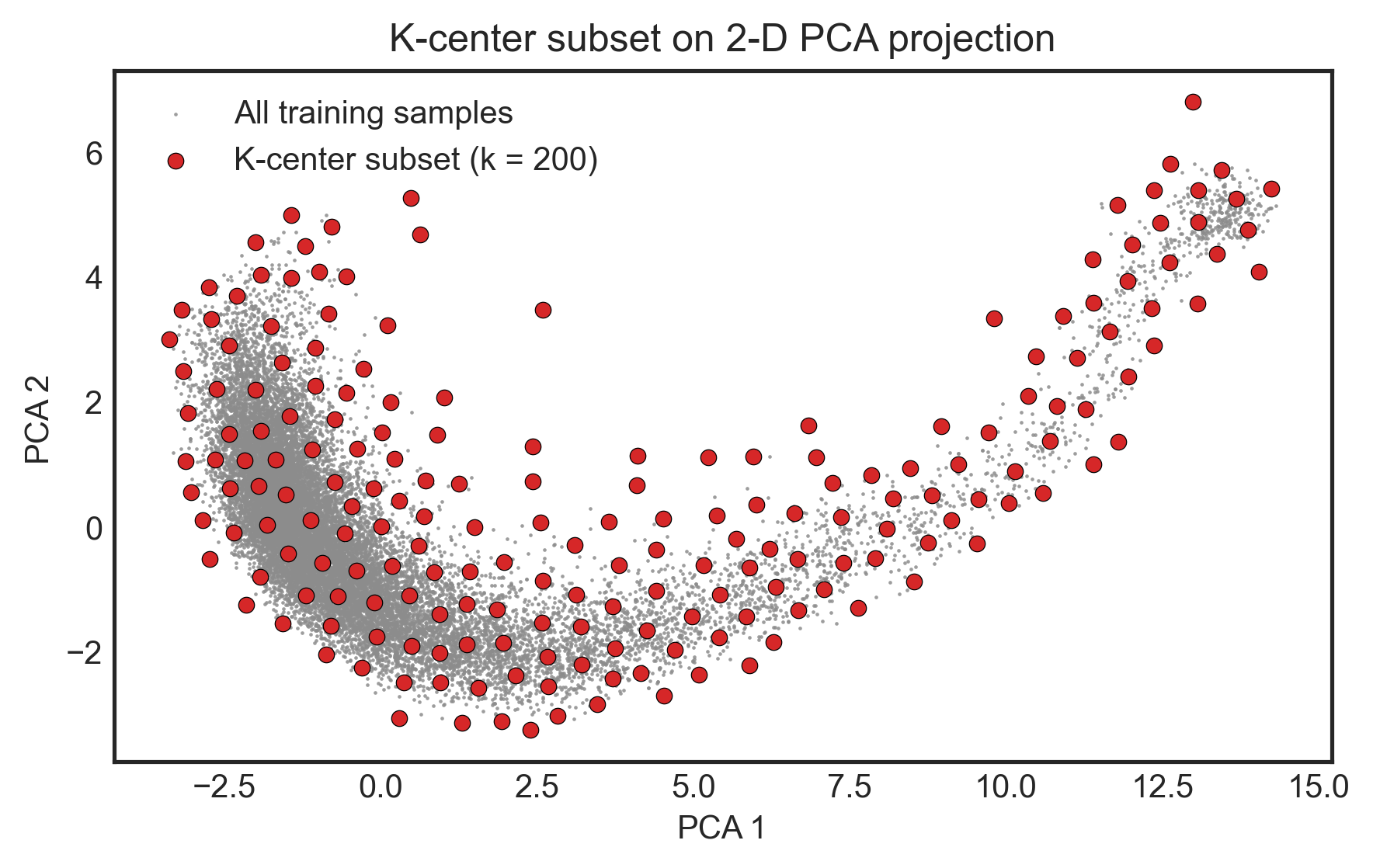}
    \caption{Visualization of $k$-center subset selection ($k=200$) from the full set of 20,000 training embeddings projected onto a 2D plane via PCA. The subset (red dots) evenly spans the embedding space, clearly illustrating the diversity-maximizing principle of the $k$-center approach. This visualization provides an intuitive geometric interpretation of how the method reduces redundancy by selecting representative points that jointly cover the entire distribution. In particular, it highlights that the selected subset is not concentrated in dense regions only, but is instead spread across both high- and low-density areas, ensuring broad coverage of the embedding space. This property is central to the effectiveness of $k$-center selection in constructing a compact yet diverse training set that preserves the structure of the full data distribution.}
    \label{fig:kcenter_pca}
\end{figure}

To do this, we formulate subset selection as an instance of the classic k-center problem. Intuitively, the k-center problem seeks to choose k points (representative embeddings, in our case) such that the maximum distance of any point in the dataset to its nearest chosen point is minimized. Figure~\ref{fig:kcenter_pca} illustrates the result of applying the k-center algorithm (with k=200) to our embeddings, which we projected into a two-dimensional space using principal component analysis (PCA). The plot shows the 200 embeddings selected by the algorithm (red) and the full 20,000-sample embedding space (gray). The selected points are spread across the embedding cloud, illustrating the diversity-maximizing behavior of the algorithm.

Formally, let $X=\{x_1,x_2,\dots,x_n\}$ be a dataset in a metric space with a distance function $d(\cdot,\cdot)$. The k-center problem can be expressed as:
$$
\min_{S\subseteq X, |S|=k} \max_{x\in X} \min_{s \in S} d(x,s)
$$
Where $S$ is the set of chosen centers, $x$ denotes a point in the dataset, and $s$ is a point in $S$. Finding an exact solution to this problem is NP-Hard \cite{hochbaum1985best}, meaning no known polynomial-time solution exists. Given the computational complexity, we implemented a greedy farthest-first procedure based on \cite{gonzalez1985clustering}. The classical 2-approximation guarantee applies to metric k-center; we do not claim the same guarantee directly for the cosine dissimilarity defined below. The greedy approach iteratively selects points, each time adding the embedding that is farthest from the current set of chosen embeddings. We provide the pseudocode below for clarity:

\begin{algorithm}[H]
    \label{alg:my_image_alg}
    \centering
    \includegraphics[width=\columnwidth]{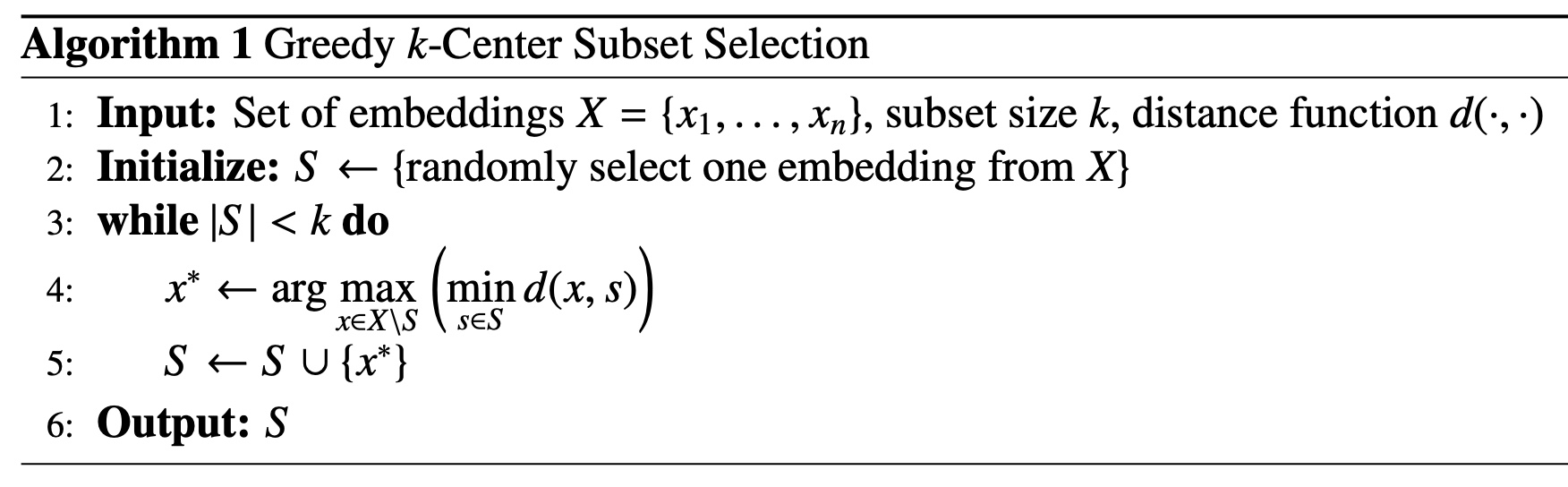}
\end{algorithm}

The computational complexity of this greedy algorithm is $O(knd)$, where $k$ is the desired subset size, $n$ is the total number of points (embeddings) in the dataset, and $d$ is the dimension of the embeddings (in our case, 768). At each iteration, the algorithm computes distances from the unselected points to the currently selected subset.

Because the embedding vectors are high-dimensional, the choice of distance metric can substantially influence the composition of the selected subset. We therefore experimented with two distance metrics: Euclidean distance and cosine distance. The Euclidean distance between two embeddings $x,y \in \mathbb{R}^{d}$ is defined as the standard geometric distance between vectors in the embedding space:
$$
d_{euc}(x,y) = \sqrt{\sum_{i=1}^d(x_i-y_i)^2}
$$
Cosine distance is derived from cosine similarity, which measures the angle between vectors:
$$
Cosine \,Similarity(x,y) =\frac{x \cdot y}{\lVert x \rVert \lVert y \rVert}, \quad d_{cos}(x,y) = 1- \frac{x \cdot y}{\lVert x \rVert \lVert y \rVert}
$$

Cosine similarity ranges from $-1$ (opposite direction) to $1$ (same direction), with $0$ indicating orthogonality (no similarity). Since cosine distance depends only on the direction, we normalize all embeddings to unit length before applying the selection algorithm.

The choice between cosine similarity and Euclidean distance significantly impacts subset selection, especially in high-dimensional spaces. In high-dimensional settings, Euclidean distances between random vectors often become less meaningful due to the "curse of dimensionality" as distances tend to concentrate around a narrow range of values \cite{franccois2007concentration}. Cosine similarity, however, measures the angle between vectors rather than their length, allowing it to distinguish between vectors even if their magnitudes differ significantly, making it often preferable for semantic embeddings spaces where direction captures meaningful differences \cite{aggarwal2001surprising}. We therefore empirically compare both metrics to see which type of geometric diversity correlates best with ASR performance.

We applied this greedy algorithm on six different subset sizes: $k \in \{ 500, 1000, 2000, 5000, 10000, 15000\}$. For each subset size $k$, we build two separate subsets: one subset using Euclidean distance and the other using cosine distance, enabling a direct comparison between these metrics. The same procedure is applied separately to both Hebrew and Danish.

\subsection{Model Fine-tuning and Evaluation}

The final stage is fine-tuning Whisper on each subset separately to evaluate the impact of our subset selection strategy. Each subset (selected by the greedy k-center algorithm using either Euclidean or cosine distances) is used independently to fine-tune the pretrained Whisper-small\footnote{https://huggingface.co/openai/whisper-small}
 checkpoint. The training process was standard: audio samples were converted into fixed-length log-Mel spectrograms by Whisper’s built-in feature extractor, and the transcripts were tokenized using Whisper’s multilingual tokenizer. During training, we monitored the WER and CER on a held-out validation set (as described previously) to measure performance and apply early stopping. Training was performed on a single NVIDIA GeForce RTX 3060 Ti with 8GB of VRAM.

In our experiments, we fine-tune all model parameters (i.e., full fine-tuning) rather than using parameter-efficient techniques such as adapters or LoRA. This choice was made to ensure a controlled and consistent evaluation protocol, allowing us to isolate the contribution of the proposed subset selection method without introducing additional variability stemming from architectural modifications or alternative training strategies. Full fine-tuning enables the model to fully adapt to the selected subsets, thereby providing a clearer signal regarding the quality and representativeness of the data selection process.

All compared selection strategies use the same fine-tuning setup. Embedding extraction and selection precede training; end-to-end computational savings and reductions in GPU utilization or peak memory were not established.

At the same time, parameter-efficient fine-tuning methods are highly relevant in low-resource scenarios, as they reduce memory footprint and computational cost by updating only a subset of the model parameters. However, these approaches introduce additional hyperparameters and design choices (e.g., adapter architecture, insertion layers, or rank selection in LoRA), which may confound the analysis of the data selection strategy itself. Moreover, the interaction between subset selection and parameter-efficient adaptation is non-trivial and may require a dedicated study. For these reasons, we leave a systematic comparison between full fine-tuning and parameter-efficient approaches as an important direction for future work.

To evaluate the fine-tuned model, we used two standard ASR metrics: WER and CER. WER measures the minimum number of word-level edits needed to transform the predicted transcription into the ground-truth reference, normalized by the number of words in the reference. Formally, given a predicted transcription and its ground truth, WER is computed as follows:
$$
WER = \frac{S+D+I}{N}
$$
where $S$ is the number of word substitutions (wrongly predicted words), $D$ is the number of deletions (words missed in prediction), $I$ is the number of insertions (extra words not present in the reference) and $N$ is the number of words in the ground truth transcription. Similarly, CER is computed at the character-level: 
$$
CER = \frac{S_c+D_c+I_c}{N_c}
$$
with $S_c$, $D_c$ and $I_c$ denoting character-level substitutions, deletions and insertions respectively, and $N_c$ is the number of characters in the reference transcription. Both WER and CER range from $0\%$ (perfect transcription) upward, lower values indicate better performance.

In this study, we focus on WER and CER as the primary evaluation metrics in order to maintain consistency with prior work in the field and to enable direct comparison with existing and future transcription systems. These metrics are widely accepted in speech recognition and transcription research due to their interpretability and strong alignment with practical transcription quality. While additional metrics such as COMET, chrF, and METEOR may provide complementary perspectives, they often rely on pretrained models or external linguistic resources, which may introduce variability, particularly in multilingual and domain-specific settings. We therefore leave the incorporation of such metrics for future work.

To establish a solid baseline and highlight the advantage of our embedding-based selection method, we also fine-tuned Whisper-small on randomly selected subsets of the same size. For each subset size (500, 1000, 2000, 5000, 10000, 15000), we independently drew three random subsets from the training pool. We fine-tuned separate Whisper models on these subsets and then averaged the resulting WER and CER across these three runs. Averaging over multiple runs ensures that the baseline comparison accounts for the variability in random subset selection, so our evaluation is more reliable and consistent. 

In addition to random sampling, we introduce a second baseline based on selecting the $k$ longest audio clips in the dataset. This heuristic reflects the intuition that longer samples may contain more linguistic content and could therefore provide stronger supervision. For each subset size, we sort the full training pool by duration and select the top $k$ longest clips. We fine-tuned Whisper on  each longest-$k$-subset using the same training settings as above. This provides an additional baseline to help determine whether diversity-based selection offers benefits beyond simply maximizing sample length.

\section{Results} \label{sec:results}

\begin{figure}[htbp]
    \centering
    \includegraphics[width=\columnwidth]{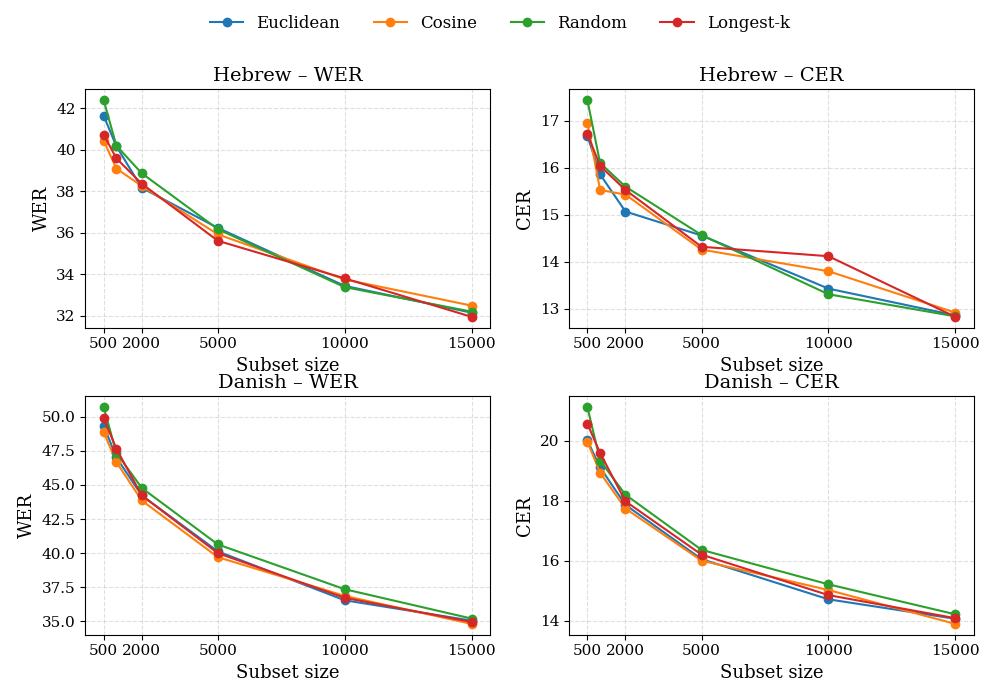} 
    \caption{CER and WER across dataset sizes for Hebrew (top) and Danish (bottom). Each subplot compares the four selection methods.}
    \label{fig:cer-wer}
\end{figure}

We evaluate the effectiveness of our subset selection strategy by fine-tuning Whisper on the subsets produced by k-center selection and comparing them against the baselines we mentioned earlier across multiple subset sizes. Table~\ref{tab:res} and Table~\ref{tab:res2} summarize the WER and CER obtained using Euclidean k-center, cosine k-center, random subsets (averaged over three runs), and the longest-k baseline. Figure~\ref{fig:cer-wer} provides a visual comparison of the WER/CER values as a function of subset size.

Across both languages, the results show a clear overall trend: k-center selection typically outperforms random sampling, particularly for smaller subset sizes. For example, with 500 samples on Hebrew, cosine k-center reduces WER from 42.39\% (random) to 40.44\% and CER from 17.44\% to 16.95\%, and similar improvements appear for Danish (from 50.69\% WER to 48.87\% WER). Although k-center does not achieve the best score in every setting, it often improves upon the random baseline, and the advantage is particularly noticeable in smaller subset sizes (500–2000 samples). As the subset size increases, all methods converge toward similar performance, and the gap between selection strategies naturally narrows.

The longest-k baseline performs better than random in several settings. Random and longest-k selection remain competitive and sometimes outperform k-center at larger sample budgets. The comparisons suggest that embedding-based selection can offer benefits beyond duration-based selection in some configurations, although it does not dominate longest-k across all settings.

The comparison between Euclidean and cosine distance reveals that no single distance metric dominates across all subset sizes. Cosine distance tends to outperform Euclidean for smaller subsets (500–2000 samples), while Euclidean occasionally matches or slightly exceeds cosine at larger subset sizes (5000–15000 samples). The results do not establish a general explanation for these differences between distance choices. Overall, the reported differences depend on the language, subset size, and evaluation metric.

\begin{table}[H]
     \caption{Comparison of WER and CER for subsets selected using Euclidean and Cosine similarity embedding-based methods versus randomly sampled subsets. Numbers in parentheses indicate relative improvements (green) or degradations (red) compared to random selection.}
  \label{tab:res}
    \centering
    \includegraphics[width=\columnwidth]{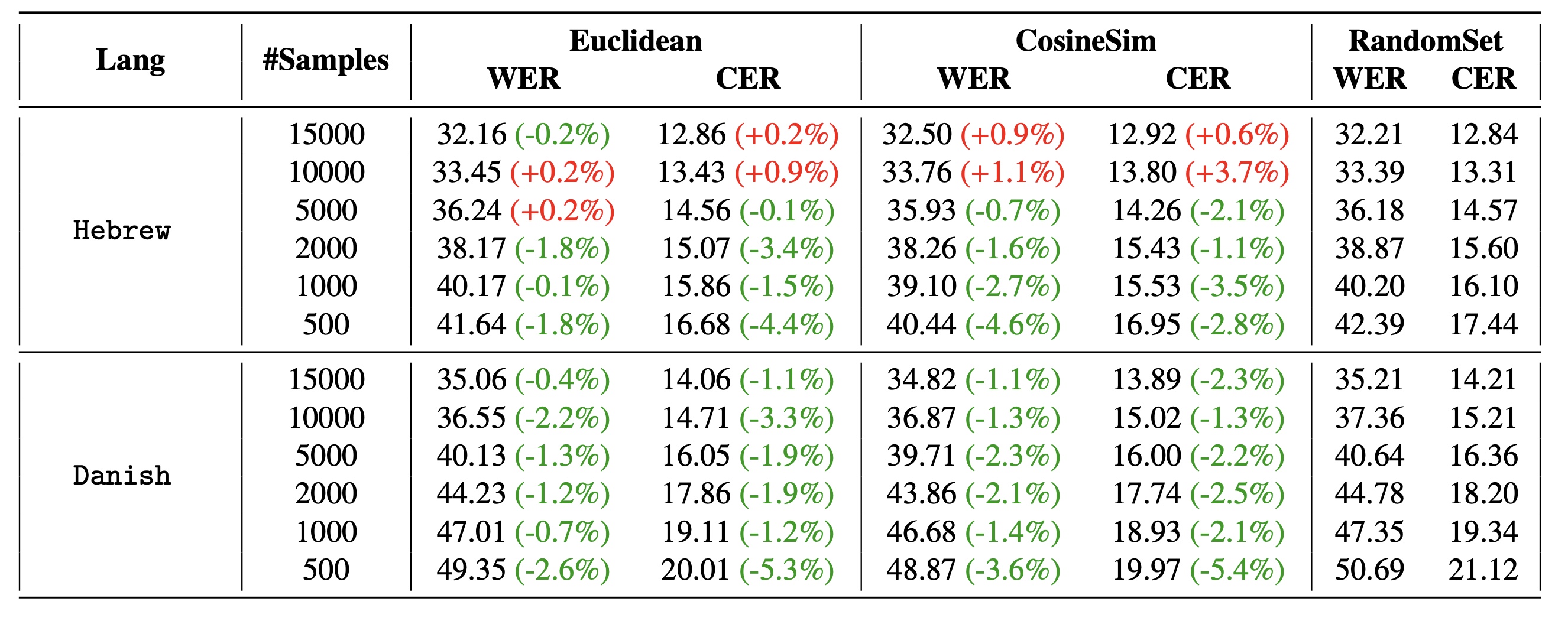}
\end{table}

\begin{table}[H]
     \caption{Comparison of WER and CER for subsets selected using Euclidean and Cosine similarity embedding-based methods versus longest-$k$-subset. Numbers in parentheses indicate relative improvements (green) or degradations (red) compared to longest-$k$-subset selection.}
  \label{tab:res2}
    \centering
    \includegraphics[width=\columnwidth]{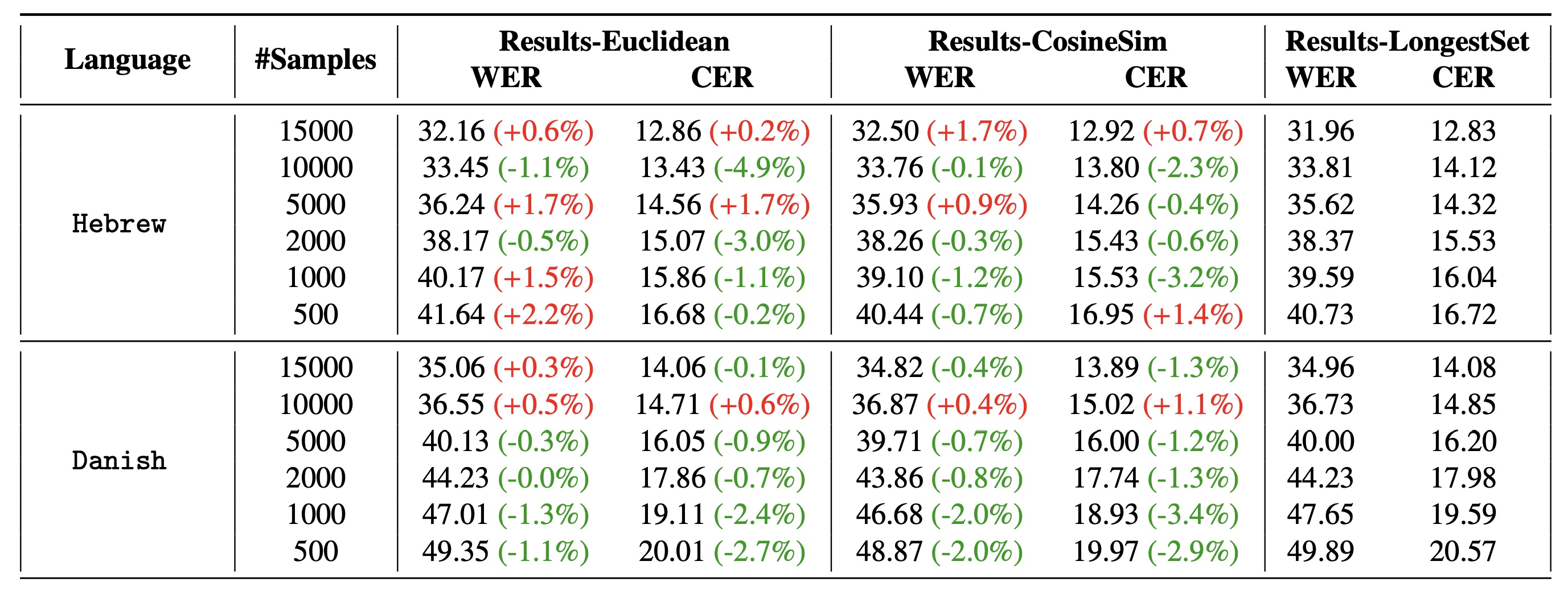}
\end{table}

\subsection{Vocabulary Coverage of Selected Subsets}
To better understand why k-center selection often leads to better ASR performance, we analyze the lexical diversity of each subset by measuring the number of unique words it contains. Higher lexical diversity indicates broader linguistic coverage, which in turn can support more effective ASR fine-tuning.

Table~\ref{tab:lexical_diversity} reports unique-word counts for cosine-based k-center subsets and random subsets. The means and standard deviations describe vocabulary counts across three random subsets, not variability in recognition performance.

\begin{table}[H]
\centering
\begin{tabular}{l c c c}
\hline
\textbf{Lang.} & \textbf{Samples} & \textbf{Rand. (mean$\pm$std)} & \textbf{k-center}\\
\hline
\multirow{4}{*}{Hebrew}
  & 20000 & -- & \textbf{38603} \\
  & 2000  & 8255.0 $\pm$ 67.7  & \textbf{13436}\\
  & 1000  & 4976.3 $\pm$ 80.1  & \textbf{10273}\\
  & 500   & 2934.3 $\pm$ 55.2  & \textbf{7404}\\
\hline
\multirow{4}{*}{Danish}
  & 20000 & -- & \textbf{40882} \\
  & 2000  & 8025.0 $\pm$ 39.0  & \textbf{8807}\\
  & 1000  & 4735.3 $\pm$ 13.7  & \textbf{5198}\\
  & 500   & 2675.0 $\pm$ 36.7  & \textbf{3046}\\
\hline
\end{tabular}
\caption{Number of unique words in each subset. Random subsets are averaged over three runs. The full 20k pool is shown for reference; its count is not a k-center selection result.\vspace{2mm}}
\label{tab:lexical_diversity}
\end{table}

For Hebrew, the evaluated k-center subsets contain more unique words than random subsets of the same size. The difference is smaller for Danish. These observations indicate greater vocabulary coverage in the reported subsets, but do not directly establish phonetic diversity or a causal explanation for recognition performance. Raw unique-word counts also depend on the amount of speech and text contained in each subset.

\subsection{Choice of Model Size}
We use Whisper-small because of the computational resources available for this study. Whisper-large was not evaluated, and published scores on different datasets do not provide a directly comparable baseline for the present results.

Our focus in this work is therefore not on achieving the lowest possible WER, but on enabling efficient and scalable adaptation under realistic constraints. This consideration is particularly important for low-resource languages, where both labeled data and computational resources are limited. In such settings, large-capacity models may be prone to overfitting, as their representational power exceeds the available data. Furthermore, the original Whisper study notes that: \textit{“also observed that many of the largest outliers in terms of worse than expected performance according to this trend are languages that have unique scripts and are more distantly related to the Indo-European languages making up the majority of the training dataset such as Hebrew (HE), Telugu (TE), Chinese (ZH), and Korean (KO).”} This highlights the additional challenges associated with adapting large pretrained models to underrepresented languages. Finally, we note that for Hebrew, no official benchmark results were reported in the original Whisper work. 

As a result, we evaluated our approach using the Ivrit.ai dataset, which provides a practical testbed for assessing model adaptation in a low-resource language setting. The results characterize recognition performance across the evaluated sample budgets; they do not establish an accuracy--cost comparison with larger models.

\subsection{Discussion}

In this work, we focus on the Whisper-small variant in order to align with the practical constraints that motivate our approach. While larger models such as Whisper-large may achieve higher absolute performance, they require substantially greater computational resources, including increased GPU memory, longer training times, and higher energy consumption, which can make them impractical in many real-world scenarios. In contrast, Whisper-small represents a more accessible and widely deployable model, suitable for environments such as edge devices, IoT settings, low-budget research groups, and rapid prototyping workflows, including training on a single GPU or even a high-end laptop.

Importantly, our goal is not only to achieve strong accuracy, but to do so under realistic efficiency constraints. By using a smaller model, we are able to better demonstrate the impact of our subset selection method in reducing data requirements while maintaining competitive performance. This setting highlights the trade-off between data efficiency and model capacity, and allows us to isolate the contribution of the data selection strategy without the confounding effects of very large models.

The offline selection procedure may be useful in repeated adaptation workflows. Any overall runtime benefit depends on preprocessing cost and the extent to which extracted embeddings can be reused.

Finally, the selection procedure could be applied when adapting other model variants, but transfer across architectures or model sizes was not evaluated. However, evaluating such models would require substantially more computational resources and is therefore left for future work.

The results are limited to Whisper-small and the sampled Hebrew and Danish corpora. Each embedding-based and longest-k configuration was evaluated with one fine-tuning run, so training variability was not quantified. Comparisons use equal numbers of utterances rather than matched audio durations, and unique-word counts provide descriptive evidence of vocabulary coverage rather than a demonstrated explanation of recognition performance.

\section{Conclusion}\label{sec:conclusion}

In this paper, we explored embedding-based subset selection for efficient fine-tuning of large pretrained ASR models. Our goal was to reduce the computational cost of adapting models like Whisper while retaining as much accuracy as possible. Using Whisper’s encoder to map each audio sample into a high-dimensional representation, we applied a greedy k-center selection strategy to identify diverse and representative subsets of the data.

Across both Hebrew and Danish, our results show that k-center selection often improves upon random sampling, particularly at smaller subset sizes. The selection stage uses a frozen encoder and does not require updating model parameters or using transcripts to rank samples; the subsequent ASR fine-tuning uses the available transcriptions. While cosine distance tends to offer an advantage at the smallest data budgets and Euclidean distance becomes competitive as subsets grow, the advantages over random sampling are not uniform across configurations.

To better understand why these subsets are effective, we also examined their linguistic diversity by measuring the number of unique words in each subset. K-center subsets contain more unique words than randomly selected subsets of the same size, with especially large gains for Hebrew and consistent, though smaller, gains for Danish. This indicates that greater geometric coverage in the embedding space corresponds to richer linguistic coverage in the selected data, which may help explain the observed improvements but does not establish their cause.

Overall, our findings demonstrate that embedding-based subset selection offers a practical approach to training-subset selection that requires no language-specific selection rules. The observed benefits are clearest at smaller sample budgets in the two evaluated languages. Training stability, overall computational savings, and applicability to other settings remain to be established.

\bibliographystyle{unsrtnat}
\bibliography{refs}  

\end{document}